\documentclass[sigchi]{acmart}

\AtBeginDocument{%
  \providecommand\BibTeX{{%
    \normalfont B\kern-0.5em{\scshape i\kern-0.25em b}\kern-0.8em\TeX}}}

\setcopyright{acmcopyright}
\copyrightyear{2020}
\acmYear{2020}
\acmDOI{10.1145/1122445.1122456}

\acmConference[CIKM '20]{CIKM '20: 29th ACM International Conference on Information and Knowledge Management}{October 19--23, 2020}{Galway, Ireland}
\acmBooktitle{CIKM '20: 29th ACM International Conference on Information and Knowledge Management, October 19--23, 2020, Galway, Ireland}
\acmPrice{15.00}
\acmISBN{978-1-4503-XXXX-X/18/06}

\def\beqrs{\begin{eqnarray*}}
\def\eeqrs{\end{eqnarray*}}
\usepackage{algorithm}
\usepackage{algorithmic}

\newcommand{\csubsection}[1]{
\begin{flushleft}
\stepcounter{subsection}
{\bf \arabic{section}.\arabic{subsection} \quad #1}
\end{flushleft}
}

\def\mS{\mathcal S}
\def\mR{\mathbb{R}}
\def\beq{\begin{equation}}
\def\eeq{\end{equation}}

\begin{document}

\title{Automatic, Dynamic, and Nearly Optimal Learning Rate Specification by Local Quadratic Approximation}

\author{Yingqiu Zhu}
\email{rozen0maiden@126.com}
\affiliation{
  \institution{School of Statistics, Renmin University of China}
  \city{Beijing}
  \country{China}
  \postcode{100872}
}
\author{Yu Chen}
\email{yu.chen@pku.edu.cn}
\affiliation{
 \institution{Guanghua School of Management, Peking University}
  \city{Beijing}
  \country{China}
}

\author{Danyang Huang}
\authornote{Corresponding Author}
\email{dyhuang@ruc.edu.cn}
\affiliation{
  \institution{Center for Applied Statistics, Renmin University of China}
  \institution{School of Statistics, Renmin University of China}
  \city{Beijing}
  \country{China}
  }

\author{Bo Zhang}
\email{mabzhang@ruc.edu.cn}
\affiliation{
  \institution{Center for Applied Statistics, Renmin University of China}
  \institution{School of Statistics, Renmin University of China}
  \city{Beijing}
  \country{China}
}

\author{Hansheng Wang}
\email{hansheng@pku.edu.cn}
\affiliation{
 \institution{Guanghua School of Management, Peking University}
  \city{Beijing}
  \country{China}
}

\renewcommand{\shortauthors}{Zhu et al.}

\begin{abstract}
In deep learning tasks, the learning rate determines the update step size in each iteration, which plays a critical role in gradient-based optimization. However, the determination of the appropriate learning rate in practice typically replies on subjective judgement. In this work, we propose a novel optimization method based on local quadratic approximation (LQA). In each update step, given the gradient direction, we locally approximate the loss function by a standard quadratic function of the learning rate. Then, we propose an approximation step to obtain a nearly optimal learning rate in a computationally efficient way. The proposed LQA method has three important features. First, the learning rate is automatically determined in each update step. Second, it is dynamically adjusted according to the current loss function value and the parameter estimates. Third, with the gradient direction fixed, the proposed method leads to nearly the greatest reduction in terms of the loss function. Extensive experiments have been conducted to prove the strengths of the proposed LQA method.

\end{abstract}




\maketitle

\section{Introduction}
In recent years, the development of deep learning has led to remarkable success in visual recognition \citep{krizhevsky2012imagenet, he2016deep, huang2017densely}, speech recognition \citep{hinton2012deep, xiong2016achieving}, natural language processing \citep{bahdanau2014neural, goldberg2014word2vec}, and many other fields. For different learning tasks, researchers have developed different network frameworks, including deep convolutional neural networks \citep{lecun1989backpropagation, krizhevsky2012imagenet}, recurrent neural networks \citep{graves2013speech}, graph convolutional networks \citep{kipf2016semi} and reinforcement learning \citep{mnih2013playing, mnih2015human}. Although the network structure could be totally different, the training methods are typically similar. They are often gradient decent methods, which are developed based on backpropagation \citep{rumelhart1995backpropagation}.

Given a differentiable objective function, gradient descent is a natural and efficient method for optimization. Among various gradient descent methods, the stochastic gradient descent (SGD) method \citep{ robbins1951stochastic} plays a critical role. In the standard SGD method, the first-order gradient of a randomly selected sample is used to iteratively update the parameter estimates  of a network. Specifically, the parameter estimates are adjusted with the negative of the random gradient multiplied by a step size. The step size is called the {\it learning rate}. Many generalized methods based on the SGD method have been proposed \citep{rumelhart1986learning,tieleman2012lecture,duchi2011adaptive,kingma2014adam,andrychowicz2016learning}. Most of these extensions specify improved update rules to adjust the direction or the step size. However, \cite{andrychowicz2016learning} pointed out that, many hand-designed update rules are designed for circumstances with certain characteristics, such as sparsity or nonconvexity. As a result, rule-based methods might perform well in some cases but poorly in others. Consequently, an optimizer with an automatically adjusted update rule is preferable.

An update rule contains two important components: one is the update direction, and the other is the step size. The learning rate determines the step size, which plays a significant role in optimization. If it is set inappropriately, the parameter estimates could be suboptimal. Empirical experience suggests that a relatively larger learning rate might be preferred in the early stages of the optimization. Otherwise, the algorithm might converge very slowly. In contrast, a relatively smaller learning rate should be used in the later stages. Otherwise, the objective function cannot be fully optimized. This phenomenon inspires us to design a method to automatically search for an optimal learning rate in each update step during optimization.

To this end, we propose here a novel optimization method based on local quadratic approximation (LQA). It tunes the learning rate in a dynamic, automatic and nearly optimal manner. The method can obtain the best step size in each update step. Intuitively, given a search direction, what should be the best step size? One natural definition is the step size that can lead to the greatest reduction in the global loss. Accordingly, the step size itself should be treated as a parameter, that needs to be optimized. For this purpose, the proposed method can be decomposed into two important steps. They are the {\it expansion} step and the {\it approximation} step. First, in the {\it expansion} step, we conduct Taylor expansion on the loss function, around the current parameter estimates. Accordingly, the objective function can be locally approximated by a quadratic function in terms of the learning rate. Then, the learning rate is also treated as a parameter to be optimized, which leads to a nearly optimal determination of the learning rate for this particular update step.

Second, to implement this idea, we need to compute the first- and second-order derivatives of the objective function on the gradient direction. One way to solve this problem is to compute the Hessian matrix for the loss function. However, this solution is computationally expensive. Because many complex deep neural networks involve a large number of parameters, this makes the Hessian matrix have ultra-high dimensionality. To solve this problem, we propose here a novel {\it approximation} step. Note that, given a fixed gradient direction, the loss function can be approximated by a standard quadratic function with the learning rate as the only input variable. For a univariate quadratic function such as this, there are only two unknown coefficients. They are the linear term coefficient and the quadratic term coefficient. As long as these two coefficients can be determined, the optimal learning rate can be obtained. To estimate the two unknown coefficients, one can try, for example, two different but reasonably small learning rates. Then, the corresponding objective function can be evaluated. This step leads to two equations, which can be solved to estimate the two unknown coefficients in the quadratic approximation function. Thereafter, the optimal learning rate can be obtained.

\textbf{Our contributions}: We propose an automatic, dynamic and nearly optimal learning rate tuning algorithm that has the following three important features.

  (1) The algorithm is automatic. In other words, it leads to an optimization method with little subjective judgment.

  (2) The method is dynamic in the sense that the learning rate used in each update step is different. It is dynamically adjusted according to the current status of the loss function and the parameter estimates. Typically, larger rates are used in the earlier iterations, while smaller rates are used in the latter iterations.

  (3) The learning rate derived from the proposed method is nearly optimal. For each update step, by the novel quadratic approximation, the learning rate leads to almost the greatest reduction in terms of the loss function. Here, ``almost'' refers to the fact that the loss function is locally approximated by a quadratic function with unknown coefficients numerically estimated. For this particular update step, with the gradient direction fixed, and among all the possible learning rates, the one determined by the proposed method can result in nearly the greatest reduction in terms of the loss function.

The rest of this article is organized as follows. In Section 2, we review related works on gradient-based optimizers. Section 3 presents the proposed algorithm in detail. In Section 4, we verify the performance of the proposed method through empirical studies on open datasets. Then, concluding remarks are given in Section 5.

\section{Related work}
To optimize a loss function, two important components need to be specified: the update direction and the step size. Ideally, the best update direction should be the gradient computed for the loss function based on the whole data. For convenience, we refer to it as the {\it global gradient}. Since the calculation of the global gradient is computationally expensive, the SGD method \citep{ robbins1951stochastic} uses the gradient estimated based on a stochastic subsample in each iteration, which we referred to as a {\it sample gradient}. It leads to a fairly satisfactory empirical performance. The SGD method has inspired many new optimization methods, most of which enhance their performance by improving the estimation of the global gradient direction. A natural improvement is to combine sample gradients from different update steps so that a more reliable estimate for the global gradient direction can be obtained. This improvement has led to the momentum-based optimization methods, such as those proposed in \cite{rumelhart1986learning,tseng1998incremental,cotter2011better,lan2012optimal}. In particular, \cite{cotter2011better} adopted Nesterov's accelerated gradient algorithm \cite{nesterov27method} to further improve the calculation of the gradient direction.

There exist other optimization methods that focus on the adjustment of the step size. \cite{duchi2011adaptive} proposed AdaGrad, in which the step size is iteratively decreased according to a prespecified function. However, it still involves a parameter related to the learning rate, which needs to be subjectively determined. More extensions of AdaGrad have been proposed, such as RMSProp \citep{tieleman2012lecture} and AdaDelta \cite{zeiler2012adadelta}. Particularly, RMSProp introduced a decay factor to adjust the weights of previous sample gradients. \citep{kingma2014adam} proposed an adaptive moment estimation (Adam) method, that combined RMSProp with a momentum-based method. Accordingly, the step size and the update direction are both adjusted during each iteration. However, because step sizes are adjusted without considering the loss function, the loss reduction obtained for each update step is suboptimal. Thus, the resulting convergence rate can be further improved.

To summarize, most existing optimization methods suffer from one or both of the following two limitations. First, they are not automatic, and human intervention is required. Second, they are suboptimal because the loss reduction achieved in each update step can be further improved. These pioneering researchers inspired us to develop a new method for automatic determination of the learning rate. Ideally, the new method should be automatic with little human intervention. It should be dynamic so that the learning rate used for each update step is particularly selected. Mostly importantly, in each update step, the learning rate determined by the new method  should be optimal (or nearly optimal) in terms of the loss reduction, given a fixed update direction.

\section{Methodology}
In this section, we first introduce the notations used in this paper and the general formulation of the SGD method. Then, we propose an algorithm based on local quadratic approximation to dynamically search for an optimal learning rate. This results in a new variant of the SGD method.

\csubsection{Stochastic gradient descent}
Assume we have a total of $N$ samples. They are indexed by $1\leq i \leq N$ and collected by $\mS=\{1,2,\cdots, N\}$. For each sample, a loss function can be defined as $\ell(X_i;\theta)$, where $X_i$ is the input corresponding to the $i$-th sample and $\theta\in\mR^p$ denotes the parameter. Then the global loss function can be defined as
$$\ell(\theta)=\frac{1}{N}\sum^N_{i=1} \ell(X_i;\theta)=\frac{1}{|\mS|}\sum_{i\in\mS}\ell(X_i;\theta).$$
Ideally, one should optimize $\ell(\theta)$ by a gradient descent algorithm. Assume there are a total of $T$ iterations. Let $\hat{\theta}^{(t)}$ be the parameter estimate obtained in the $t$-th iteration. Then, the estimate in the next iteration $\hat{\theta}^{(t+1)}$ is given by,
\begin{equation*}
  \hat{\theta}^{(t+1)}= \hat{\theta}^{(t)}-\delta \nabla \ell(\hat{\theta}^{(t)}),
\label{GD}
\end{equation*}
where $\delta$ is the learning rate and $\nabla \ell(\hat{\theta}^{(t)})$ is the gradient of the global loss function $\ell(\theta)$  with respect to $\theta$ at $\hat{\theta}^{(t)}$. More specifically, $\nabla \ell(\hat{\theta}^{(t)})=N^{-1}\sum_{i=1}^N \nabla \ell(X_i;\hat{\theta}^{(t)})$, where $\nabla \ell(X_i;\hat{\theta}^{(t)})$ is the gradient of the local loss function for the $i$-th sample.

Unfortunately, such a straightforward implementation is computationally expensive if the sample size $N$ is relatively large, which is particularly true if the dimensionality of $\theta$ is also ultrahigh. To alleviate the computational burden, researchers proposed the idea of SGD. The key idea is to randomly partition the whole sample into a number of nonoverlapping batches. For example, we can write $\mS=\cup^K_{k=1} \mS_k$, where $\mS_k$ collects the indices of the samples in the $k$-th batch. We should have $\mS_{k_1} \cap \mS_{k_2} = \emptyset$ for any $k_1 \neq k_2$ and $|\mS_k|=n$ for any $1\leq k \leq K$, where $n$ is a fixed batch size. Next, instead of computing the global gradient $\nabla \ell(\hat{\theta}^{(t)})$, we can replace it by an estimate computed based on the $k$-th batch. More specifically, each iteration (e.g., the $t$-th iteration) is further decomposed into a total of $K$ batch steps. Let $\hat{\theta}^{(t,k)}$ be the estimate obtained during the $k$-th ($1\leq k\leq K$) batch step during the $t$-th iteration. Then, we have
\begin{equation*}
  \hat{\theta}^{(t, k+1)} = \hat{\theta}^{(t, k)}- \frac{\delta}{n} \sum_{i \in \mS_k} \nabla \ell(X_i;\hat{\theta}^{(t,k)}),
\end{equation*}
where $k=1,\cdots, K-1$. In particular, $\hat{\theta}^{(t+1, 1)} = \hat{\theta}^{(t, K)}-\delta n^{-1}\sum_{i \in \mS_K} \nabla \ell(X_i;\hat{\theta}^{(t,K)})$.

By doing so, the computational burden can be alleviated. However, the tradeoff is that the batch-sample-based gradient estimate could be unstable, which is particularly true if the batch size $n$ is relatively small. To fix this problem, various momentum-based methods have been proposed. The key idea is to record gradients in previous iterations and integrate them together to form a more stable estimate.

\csubsection{Local quadratic approximation}
In this work, we assume that for each batch step, the estimate for the gradient direction is given. It can be obtained by different algorithms. For example, it could be the estimate obtained by a standard SGD algorithm or an estimate that involves rule-based corrections, such as that from a momentum-based method. We focus on how to specify the learning rate in an optimal (or, nearly optimal) way.

 To this end, we treat the learning rate $\delta$ as an unknown parameter. It is remarkable that the optimal learning rate could dynamically change in different batch steps. Thus, we use $\delta_{t, k}$ to denote the learning rate in the $k$-th batch step within the $t$-th iteration. Since the reduction in the loss in this batch step is influenced by the learning rate $\delta_{t, k}$, we express it as a function of the learning rate $\Delta \ell(\delta_{t, k})$.

To find the optimal value for $\delta_{t, k}$, we investigate the optimization of $\Delta \ell(\delta_{t, k})$ based on the Taylor expansion. For simplicity, we use $g_{t,k} = n^{-1}\sum_{i \in \mS_k} \nabla \ell(X_i; \hat{\theta}^{(t,k)})$ to denote the current gradient. Given $\hat{\theta}^{(t, k)}$ and $g_{t,k}$, the loss reduction could be expressed as
\begin{align*}
\Delta \ell(\delta_{t, k}) &= \frac{1}{n}\sum_{i \in \mS_k}  \left[\ell \left(X_i;\hat{\theta}^{(t,k+1)}\right) - \ell \left(X_i;\hat{\theta}^{(t,k)}\right)\right] \\
&= \frac{1}{n}\sum_{i \in \mS_k}  \left[\ell \left(X_i;\hat{\theta}^{(t,k)}-\delta_{t, k} g_{t,k}\right) - \ell \left(X_i;\hat{\theta}^{(t,k)}\right)\right]
\end{align*}
Then, two estimation steps are conducted to determine an appropriate value for $\delta_{t,k}$ in this batch step.

\noindent
\textbf{(1) Expansion Step.} By a Taylor expansion of $\ell(X_i;\theta)$ around $\hat{\theta}^{(t,k)}$, we have $\ell \left(X_i;\hat{\theta}^{(t,k)} -\delta_{t,k} g_{t,k}\right) =$
\begin{align*}
 & \ell \left(X_i;\hat{\theta}^{(t,k)}\right)- \nabla \ell \left(X_i;\hat{\theta}^{(t,k)}\right)\delta_{t,k} g_{t,k} \\
&~~~~~~~~~~~~~~~~~~~~~~+ \frac{1}{2} \delta^2_{t,k} g_{t,k}^T \nabla^2\ell \left(X_i;\hat{\theta}^{(t,k)}\right)g_{t,k} + o \left(\delta^2_{t,k} g_{t,k}^T g_{t,k} \right),
\end{align*}
where $\nabla \ell \left( X_i;\hat{\theta}^{(t,k)} \right)$ and $\nabla^2 \ell \left( X_i;\hat{\theta}^{(t,k)} \right)$ denote the first- and second-order derivatives of the local loss function, respectively. As a result, the reduction is
\begin{align}
&\Delta \ell(\delta_{t,k}) = \frac{1}{n}\sum_{i \in \mS_k} \left[ - \nabla \ell \left( X_i;\hat{\theta}^{(t,k)} \right) \delta_{t,k} g_{t,k} \right.
 \nonumber\\
&+\frac{1}{2} \delta^2_{t,k} g_{t,k}^T \nabla^2 \ell \left( X_i;\hat{\theta}^{(t,k)} \right) g_{t,k} + \left. o \left( \delta^2_{t,k} g_{t,k}^T g_{t,k} \right)  \right] \nonumber\\
=& - \left\{ \frac{1}{n} \sum_{i \in \mS_k} \nabla \ell \left( X_i;\hat{\theta}^{(t,k)} \right) g_{t,k} \right\} \delta_{t,k} \nonumber\\
&+ \left\{ \frac{1}{2n} \sum_{i \in \mS_k}  g_{t,k}^T \nabla^2\ell \left( X_i;\hat{\theta}^{(t,k)} \right)g_{t,k} \right\} \delta^2_{t,k} \nonumber\\
&+ o \left( n^{-1}\delta^2_{t,k} g_{t,k}^T g_{t,k} \right).
\label{Taylor_expansion}
\end{align}

According to (\ref{Taylor_expansion}), $\Delta \ell(\delta_{t,k})$ is a quadratic function of $\delta_{t,k}$. For simplicity, the coefficient of the linear term and the coefficient of the quadratic term are denoted as
\beqrs
a_{t,k} &=& \frac{1}{n}\left\{ \sum_{i \in \mS_k} \nabla \ell \left(X_i;\hat{\theta}^{(t,k)}\right)g_{t,k} \right\}, \mbox{~~~~and}\\
 b_{t,k} &=& \left\{ \frac{1}{2n} \sum_{i \in \mS_k}  g_{t,k}^T \nabla^2\ell \left(X_i;\hat{\theta}^{(t,k)}\right)g_{t,k} \right\},
 \eeqrs
  respectively. Since the Taylor remainder here could be negligible, (\ref{Taylor_expansion}) can be simply denoted by
\begin{equation}
    \Delta \ell (\delta_{t,k}) \approx -a_{t,k} \delta_{t,k} + b_{t,k} \delta^2_{t,k}.
    \label{simple}
\end{equation}
To maximize $\Delta \ell(\delta_{t,k})$ with respect to $\delta_{t,k}$, we take the corresponding derivative of the loss reduction, which leads to,
\begin{equation*}
\frac{\partial \Delta \ell(\delta_{t,k})}{\partial \delta_{t,k}} \approx -a_{t,k} + 2b_{t,k} \delta_{t,k} = 0.
\end{equation*}
As a result, the optimal learning rate in this batch step can be approximated by,
 \beq\label{eq:delta}
 \delta^*_{t,k} = (2b_{t,k})^{-1}a_{t,k}.
 \eeq

Note that the computation of $b_{t,k}$ involves
the first- and second-order derivatives. For a general form of loss function, this calculation may be computationally expensive in real applications. Thus, an approximation step is preferred to improve the computational efficiency.

\noindent \textbf{(2) Approximation Step.} To compute the coefficients $a_{t,k}$ and $b_{t,k}$ and avoid the computation of second derivatives, we consider the following approximation method. The basic idea is to build 2 equations with respect to the 2 unknown coefficients.

Let $g_{t,k}$ be a given estimate of the gradient direction. We then compute
 \begin{align}
\sum_{i \in \mS_k} \ell \left(X_i; \hat{\theta}^{(t,k)} - \delta_0 g_{t,k}\right) =& \sum_{i \in \mS_k} \ell \left(X_i; \hat{\theta}^{(t,k)}\right) - a_{t,k}\delta_0 n \nonumber\\
&+ b_{t,k}\delta_0^2 n, \label{ap_eq1}\\
\sum_{i \in \mS_k} \ell \left(X_i; \hat{\theta}^{(t,k)} + \delta_0 g_{t,k}\right) =& \sum_{i \in \mS_k} \ell \left(X_i; \hat{\theta}^{(t,k)}\right) + a_{t,k}\delta_0 n \nonumber\\
&+ b_{t,k}\delta_0^2 n, \label{ap_eq2}
\end{align}

\noindent
for a reasonably small learning rate $\delta_0$. A natural choice for $\delta_0$ could be $\delta^*_{t,k-1}$ if $k>1$ and $\delta^*_{t-1,K}$ if $k=1$. By solving (\ref{ap_eq1}) and (\ref{ap_eq2}), we have
\begin{align}
\tilde{b}_{t,k} =& \frac{1}{2n \delta^2_0} \sum_{i \in \mS_k} \left[\ell \left(X_i; \hat{\theta}^{(t,k)} + \delta_0 g_{t,k}\right) \right. \nonumber\\
&+ \left. \ell \left(X_i; \hat{\theta}^{(t,k)} - \delta_0 g_{t,k}\right) - 2  \ell \left(X_i;  \hat{\theta}^{(t,k)}\right) \right], \label{cpa}\\
\tilde{a}_{t,k} =& \frac{1}{{2n \delta_0}} \sum_{i \in \mS_k} \left[\ell \left(X_i; \hat{\theta}^{(t,k)} + \delta_0 g_{t,k}\right) \right. \nonumber\\
&- \left. \ell \left(X_i; \hat{\theta}^{(t,k)} - \delta_0 g_{t,k}\right) \right], \label{cpb}
\end{align}
where $\tilde{a}_{t,k}$ and $\tilde{b}_{t,k}$ could serve as the approximations of $a_{t,k}$ and $b_{t,k}$, respectively. Then, we apply these results back to (\ref{eq:delta}), which gives the approximated optimal learning rate $\hat{\delta}^*_{t,k}$. Because $\hat{\delta}^*_{t,k}$ is optimally selected, the reduction in the loss function is nearly optimal for each batch step. As a consequence, the total number of  iterations required for convergence can be much reduced, which makes the whole algorithm converge much faster than usual. In summary, \textbf{Algorithm \ref{alg}} illustrates the pseudocode of the proposed method.

\begin{algorithm}[h]
\caption{Local quadratic approximation algorithm}
\begin{algorithmic}\label{alg}
\STATE \textbf{Require}: $T$: number of iterations; $K$: number of batches within one iteration; $\ell(\theta)$: loss function with parameters $\theta$; $\theta_0$: initial estimate for parameters (e.g., a zero vector); $\delta_0$: initial (small) learning rate. \
\STATE $t \leftarrow 1$; \
\STATE $\hat{\theta}^{(1,1)} \leftarrow \theta_0$; \
\WHILE{$t \leq T$}
  \STATE $k \leftarrow 1$; \
  \WHILE{$k \leq K-1$}
    \STATE Compute the gradient $g_{t,k}$; \
    \STATE Compute $\tilde{a}_{t,k}$ and $\tilde{b}_{t,k}$ according to (\ref{cpa}) and (\ref{cpb}); \
    \STATE $\hat{\delta}^*_{t,k} \leftarrow (2\tilde{b}_{t,k})^{-1}\tilde{a}_{t,k}$; \
    \STATE $\hat{\theta}^{(t,k+1)} \leftarrow \hat{\theta}^{(t,k)}-\hat{\delta}^*_{t,k} g_{t,k}$; \
    \STATE $k \leftarrow k+1$; \
\ENDWHILE
\STATE Compute the gradient $g_{t,K}$; \
\STATE Compute $\tilde{a}_{t,K}$ and $\tilde{b}_{t,K}$ according to (\ref{cpa}) and (\ref{cpb}); \
\STATE $\hat{\delta}^*_{t,K} \leftarrow \left(2\tilde{b}_{t,K}\right)^{-1}\tilde{a}_{t,K}$; \
\STATE $\hat{\theta}^{(t+1,1)} \leftarrow \hat{\theta}^{(t,K)}-\hat{\delta}^*_{t,K} g_{t,K}$; \
\STATE $t \leftarrow t+1$; \
\ENDWHILE
\RETURN{ $\hat{\theta}^{(T+1,1)}$, the resulting estimate. }
\end{algorithmic}
\end{algorithm}

It is remarkable that the computational cost required for calculating the optimal learning rate is ignorable. The main cost is due to the calculation of the loss function values (not its derivatives) at two different points. The cost of this step is substantially smaller than computing the gradient and it is particularly true if the unknown parameter $\theta$'s dimension is ultrahigh.

\section{Experiments}
In this section, we empirically evaluate the proposed method based on different models and compare it with various optimizers under different parameter settings. The details are listed as follows. \\
 \noindent
 \textbf{Classification Model.} To demonstrate the robustness of the proposed method, we consider three classic models. They are multinomial logistic regression, multilayer perceptron (MLP), and deep convolutional neural network (CNN) models.\\
 \noindent
 \textbf{Competing Optimizers}. For comparison purposes, we compare the proposed LQA method with other popular optimizers. They are the standard SGD method, the SGD method with momentum, the SGD method based on Nesterov's accelerated gradient (NAG), AdaGrad, RMSProp and Adam. For simplicity, we use ``SGD-M'' to denote the SGD method with momentum and ``SGD-NAG'' to denote the SGD method based on the NAG algorithm.\\
 \noindent
 \textbf{Parameter Settings.} For the competing optimizers, we adopt three different learning rates, $\delta=0.1,0.01,$ and $0.001$. For all the optimizers, the minibatch size is 64, and the initial values for all the parameters are zero. If there are other hyperparameters (e.g., decay rates) in the models, they are set by default.\\
 \noindent
 \textbf{Performance Measurement.} To gauge the performance of the optimizers, we report the {\it training loss} of the different optimizers, which is defined as the negative log-likelihood function. The results of different optimizers in each iteration are shown in figures for comparison purposes.

\begin{figure*}[h]
  \centering
  \includegraphics[width=\linewidth]{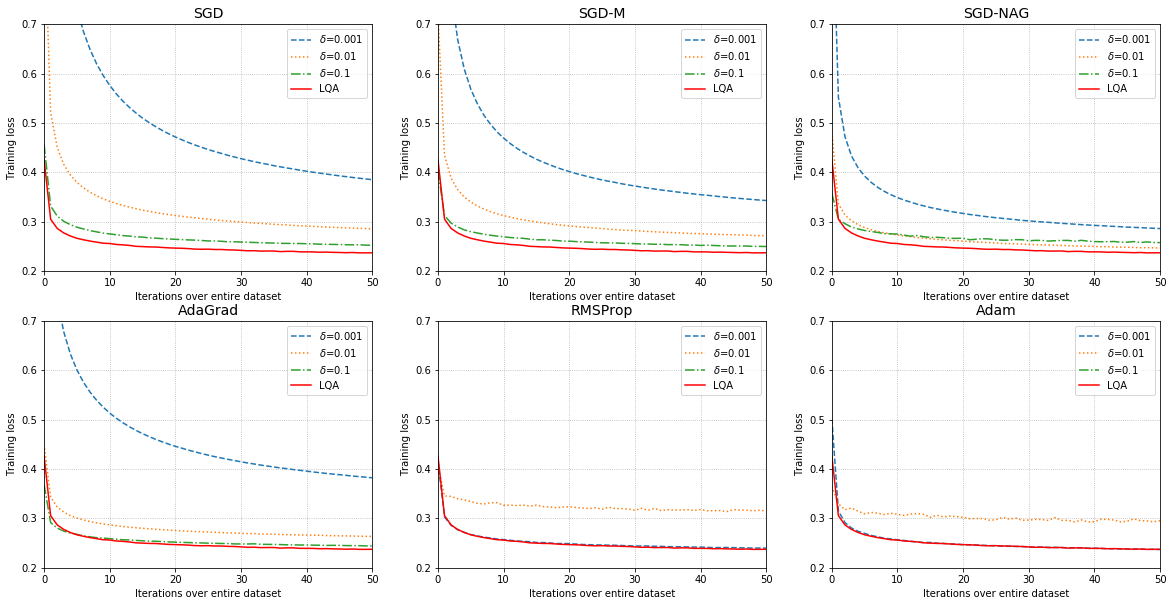}
  \caption{Training loss of multinomial logistic regression on the MNIST dataset.}
  \label{lr}
\end{figure*}

\csubsection{Multinomial Logistic Regression}
We first compare the performance of different optimizers based on multinomial logistic regression. It has a convex objective function. We consider the MNIST dataset \citep{lecun1998gradient} for illustration. The dataset consists of a total of 70,000 (28$\times$28) images of handwritten digits, each of which corresponds to a 10-dimensional one-hot vector as its label. Then, the images are flattened as 784-dimensional vectors to train the logistic regression classifier.  Figure \ref{lr} displays the performance of the proposed LQA method against the competing optimizers. We can draw the following conclusions.

\textbf{Learning Rate.} For the competing optimizers, the training loss curves of different learning rates clearly have different shapes, which means different convergence speeds because the convergence speed is greatly affected by $\delta$. The best $\delta$ in this case is 0.01 for the SGD, SGD-M, SGD-NAG and AdaGrad methods. However, that for RMSProp and Adam is 0.001. Note that for RMSProp and Adam, the loss may fail to converge with an inappropriate $\delta$ (e.g., $\delta=0.1$). It is remarkable that with the LQA method, the learning rate is automatically determined and dynamically updated. Thus, the proposed method can provide an automatic solution for the training of multinomial logistic regression classifiers. Next, we compare LQA and the competing optimizers with their best learning rates.

\textbf{Loss Reduction.} First, the loss curve of LQA remains lower than those of the SGD optimizers during the whole training process. This finding means that LQA converges faster than the SGD optimizers. For example, LQA reduces the loss to 0.256 in the first 10 iterations. It takes 40 iterations for the standard SGD optimizer with $\delta=0.1$ to achieve the same level. Second, for AdaGrad and RMSProp with $\delta=0.1$, although the training loss curves are slightly lower than that of LQA in the early stages (e.g. the first 5 iterations), LQA performs better in the later stages. Third, the best performance for Adam in this case is achieved when $\delta=0.001$. Although the performances are quite similar for LQA and Adam with $\delta=0.001$, LQA has lower loss values in the early stages.

\csubsection{Multilayer Perception}
MLP models are powerful neural network models and have been widely used in machine learning tasks \citep{ramchoun2016multilayer}. They contain multiple fully connected layers and activation functions between those layers. An MLP can approximate arbitrary continuous functions over compact input sets \citep{hornik1991approximation}.

\begin{figure*}[h]
  \centering
  \includegraphics[width=\linewidth]{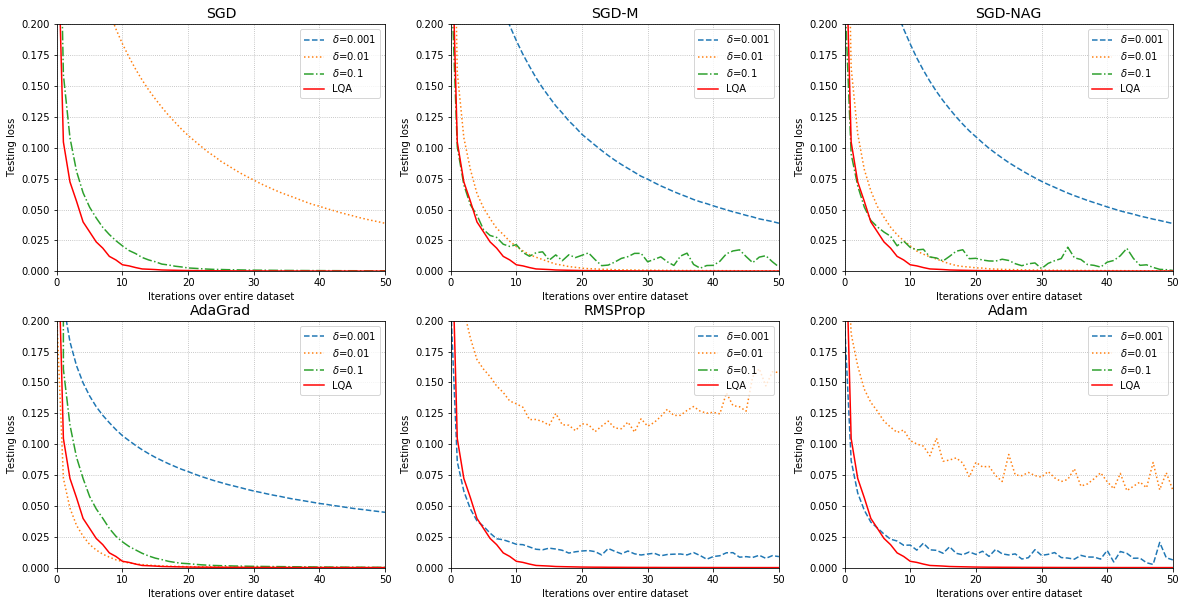}
  \caption{Training loss of MLP on the MNIST dataset. }
  \label{mlp}
\end{figure*}

To investigate the performance of the proposed method in this case, we consider the MNIST dataset. Following the model setting in \cite{kingma2014adam}, the MLP is built with 2 fully connected hidden layers, each of which has 1,000 units, and the ReLU function is adopted as the activation function. Figure \ref{mlp} shows the performances of different optimizers. The following conclusions can be drawn.

\textbf{Learning Rate.} In this case, the best learning rates for the competing methods are quite different: (1) for the standard SGD method, the best learning rate is $\delta=0.1$; (2) for the SGD-M, SGD-NAG and AdaGrad methods, the best learning rate is 0.01; (3) for RMSProp and Adam, $\delta=0.001$ is the best. It is remarkable that even for the same optimizer, different learning rates could lead to a different performance, if the model changes. Thus, determining the appropriate learning rate in practice may depend on expert experience and subjective judgement. In contrast, the proposed method can avoid such effort in choosing $\delta$ and give a comparable and robust performance.

\textbf{Loss Reduction.} First, compared with the standard SGD optimizers, LQA performs much better, which could be seen from the lower training loss curve. Second, compared with the SGD-M, SGD-NAG, RMSProp and Adam optimizers, the performance of LQA is comparable to their best performances in the early stages (e.g., the first 5  iterations). In the later stages, the LQA method continues to reduce the loss, which makes the training loss curve of LQA lower than that of the other methods. For example, the smallest loss corresponding to the Adam optimizer in the 20th iteration is 0.011, while that of the LQA method is 0.002. Third, although AdaGrad converges faster when $\delta=$0.01, the performance of the proposed method is slightly better than AdaGrad after the 12th iteration.

\csubsection{Deep Convolutional Neural Networks}
CNNs have brought remarkable breakthroughs in computer vision tasks over the past two decades \citep{krizhevsky2012imagenet, he2016deep, huang2017densely} and play a critical role in various industrial applications, such as face recognition \citep{wen2016discriminative} and driverless vehicles \citep{li2019stereo}. In this subsection, we investigate the performance of the LQA method with respect to the training of CNNs. Two classic CNNs are considered. They are LeNet \citep{lecun1998gradient} and ResNet \citep{he2016deep}. More specifically, LeNet-5 and ResNet-18 are studied in this paper. The MNIST and CIFAR10 \citep{krizhevsky2009learning} datasets are used to demonstrate the performance. The CIFAR10 dataset contains 60,000 (32$\times$32) RGB images, which are divided into 10 classes.

\begin{figure*}[h]
  \centering
  \includegraphics[width=\linewidth]{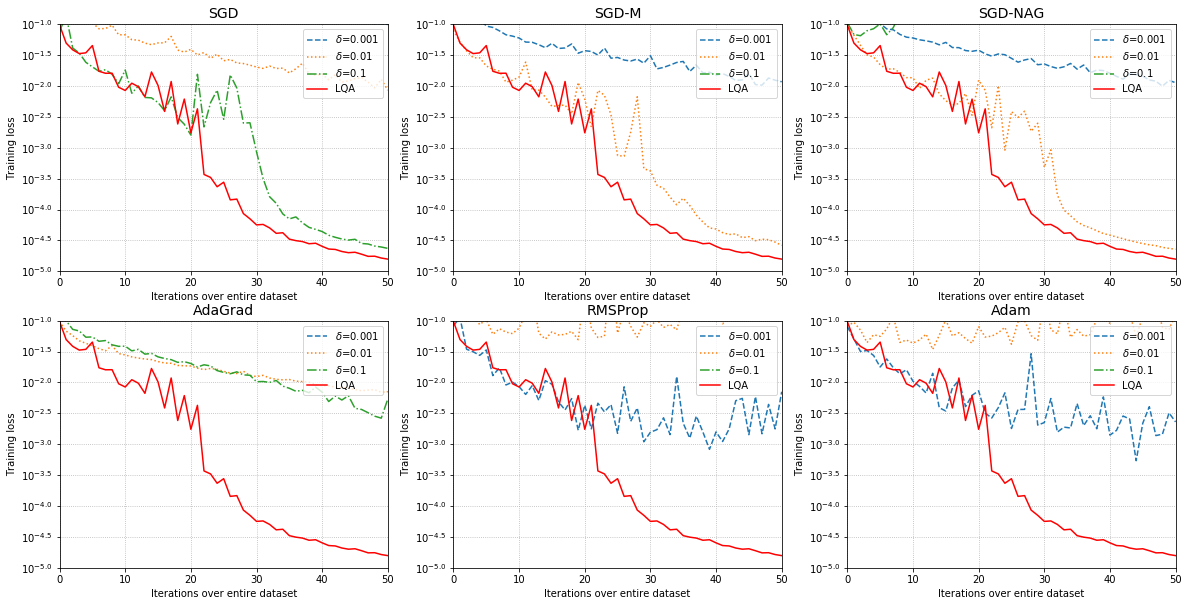}
  \caption{Training loss of LeNet-5 on the MNIST dataset. }
  \label{lenet-mnist}
\end{figure*}

\begin{figure*}[h]
  \centering
  \includegraphics[width=\linewidth]{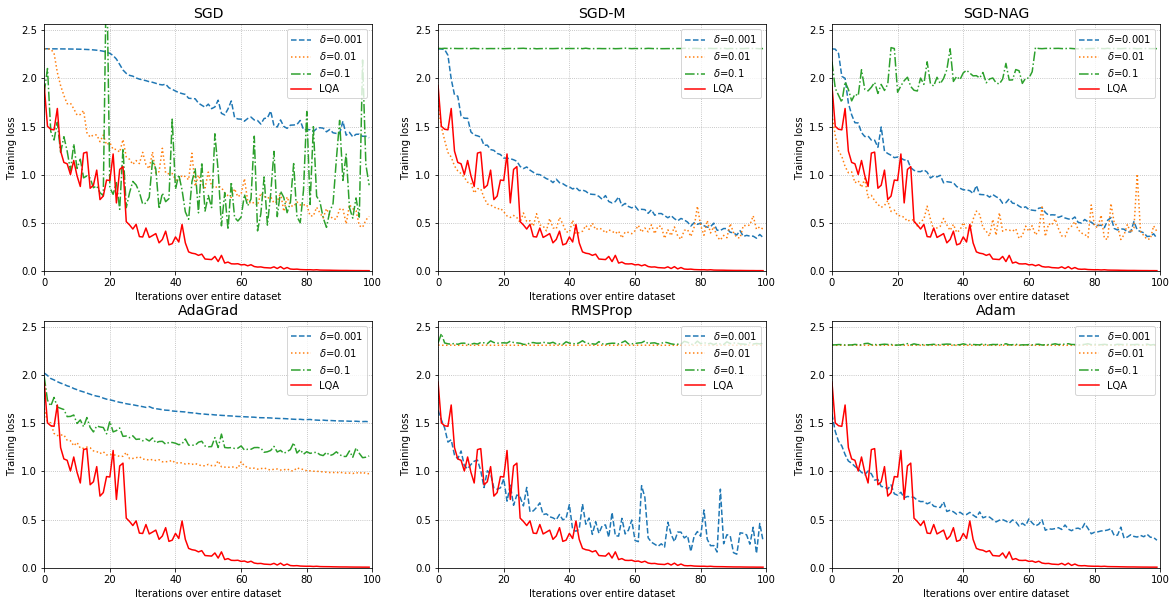}
  \caption{Training loss of LeNet-5 on the CIFAR10 dataset. }
  \label{lenet-cifar10}
\end{figure*}

\textbf{LeNet.} Figure \ref{lenet-mnist} and Figure \ref{lenet-cifar10} show the results of experiments on the MNIST and the CIFAR10 datasets, respectively. The following conclusions can be drawn: (1) For both datasets, the loss curves of the LQA method remain lower than those of the standard SGD and AdaGrad optimizers. This finding suggests LQA converges faster than those optimizers during the whole training process. (2) LQA performs similarly to the SGD-M, SGD-NAG, RMSProp and Adam optimizers in the early stages (e.g., the first 20 iterations). However, in the later stages, the proposed method can further reduce the loss and lead to a lower loss than those optimizers after the same number of iterations. (3) For the CIFAR10 dataset, a large $\delta$ (e.g., $\delta=0.1$) may lead to an unstable loss curve for a standard SGD optimizer. Although the loss curve of LQA is unstable in the early stages of training, it becomes smooth in the later stages because the proposed method is able to automatically and adaptively adjust the update step size to accelerate training. It is fairly robust.

\begin{figure*}[h]
  \centering
  \includegraphics[width=\linewidth]{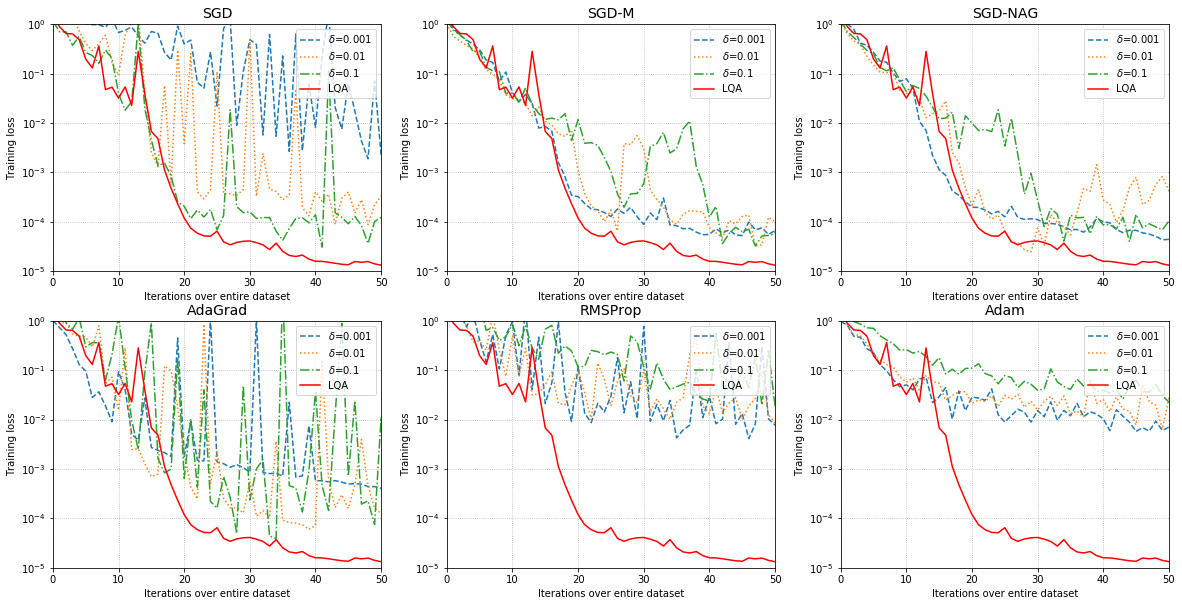}
  \caption{Training loss of ResNet-18 on the CIFAR10 dataset. }
  \label{resnet-cifar10}
\end{figure*}

\textbf{ResNet.} Figure \ref{resnet-cifar10} displays the training loss of ResNet-18 corresponding to different optimizers on the CIFAR10 dataset. Accordingly, we make the following conclusions. First, the LQA method performs similarly to the other optimizers in the early stages of training (e.g., the first 15 iterations). However, it converges faster in the later stages. Particularly, the proposed method leads to a lower loss than RMSProp and Adam within the same number of iterations. Second, in this case, the loss curves of the SGD optimizers and AdaGrad are quite unstable during the whole training period. The LQA method is much more stable in the later stages of the training than early stages.

\section{Conclusions}
In this work, we propose LQA, a novel approach to determine the nearly optimal learning rate for automatic optimization. Our method has three important features. First, the learning rate is automatically estimated in each update step. Second, it is dynamically adjusted during the whole training process. Third, given the gradient direction, the learning rate leads to nearly the greatest reduction in the loss function. Experiments on openly available datasets demonstrate its effectiveness.

We discuss two interesting topics for future research. First, the optimal learning rate derived by LQA is shared by all dimensions of the parameter estimate. A potential extension is to allow for different optimal learning rates for different dimensions. Second, in this paper, we focus on accelerating the training of the network models. We do not discuss the overfitting issue and sparsity of the gradients. To further improve the performance of the proposed method, it is possible to combine dropouts or sparsity penalties with LQA.

\bibliographystyle{ACM-Reference-Format}
%\bibliography{sample-base}

\begin{thebibliography}{30}

%%% the Web address.

\ifx \showCODEN    \undefined \def \showCODEN     #1{\unskip}     \fi
\ifx \showDOI      \undefined \def \showDOI       #1{#1}\fi
\ifx \showISBNx    \undefined \def \showISBNx     #1{\unskip}     \fi
\ifx \showISBNxiii \undefined \def \showISBNxiii  #1{\unskip}     \fi
\ifx \showISSN     \undefined \def \showISSN      #1{\unskip}     \fi
\ifx \showLCCN     \undefined \def \showLCCN      #1{\unskip}     \fi
\ifx \shownote     \undefined \def \shownote      #1{#1}          \fi
\ifx \showarticletitle \undefined \def \showarticletitle #1{#1}   \fi
\ifx \showURL      \undefined \def \showURL       {\relax}        \fi
% invisible to TeX
\providecommand\bibfield[2]{#2}
\providecommand\bibinfo[2]{#2}
\providecommand\natexlab[1]{#1}
\providecommand\showeprint[2][]{arXiv:#2}

\bibitem[\protect\citeauthoryear{Andrychowicz, Denil, Gomez, Hoffman, Pfau,
  Schaul, Shillingford, and De~Freitas}{Andrychowicz et~al\mbox{.}}{2016}]%
        {andrychowicz2016learning}
\bibfield{author}{\bibinfo{person}{Marcin Andrychowicz}, \bibinfo{person}{Misha
  Denil}, \bibinfo{person}{Sergio Gomez}, \bibinfo{person}{Matthew~W Hoffman},
  \bibinfo{person}{David Pfau}, \bibinfo{person}{Tom Schaul},
  \bibinfo{person}{Brendan Shillingford}, {and} \bibinfo{person}{Nando
  De~Freitas}.} \bibinfo{year}{2016}\natexlab{}.
\newblock \showarticletitle{Learning to learn by gradient descent by gradient
  descent}. In \bibinfo{booktitle}{\emph{Advances in neural information
  processing systems}}. \bibinfo{pages}{3981--3989}.
\newblock


\bibitem[\protect\citeauthoryear{Bahdanau, Cho, and Bengio}{Bahdanau
  et~al\mbox{.}}{2014}]%
        {bahdanau2014neural}
\bibfield{author}{\bibinfo{person}{Dzmitry Bahdanau},
  \bibinfo{person}{Kyunghyun Cho}, {and} \bibinfo{person}{Yoshua Bengio}.}
  \bibinfo{year}{2014}\natexlab{}.
\newblock \showarticletitle{Neural machine translation by jointly learning to
  align and translate}.
\newblock \bibinfo{journal}{\emph{arXiv preprint arXiv:1409.0473}}
  (\bibinfo{year}{2014}).
\newblock


\bibitem[\protect\citeauthoryear{Cotter, Shamir, Srebro, and Sridharan}{Cotter
  et~al\mbox{.}}{2011}]%
        {cotter2011better}
\bibfield{author}{\bibinfo{person}{Andrew Cotter}, \bibinfo{person}{Ohad
  Shamir}, \bibinfo{person}{Nati Srebro}, {and} \bibinfo{person}{Karthik
  Sridharan}.} \bibinfo{year}{2011}\natexlab{}.
\newblock \showarticletitle{Better mini-batch algorithms via accelerated
  gradient methods}. In \bibinfo{booktitle}{\emph{Advances in neural
  information processing systems}}. \bibinfo{pages}{1647--1655}.
\newblock


\bibitem[\protect\citeauthoryear{Duchi, Hazan, and Singer}{Duchi
  et~al\mbox{.}}{2011}]%
        {duchi2011adaptive}
\bibfield{author}{\bibinfo{person}{John Duchi}, \bibinfo{person}{Elad Hazan},
  {and} \bibinfo{person}{Yoram Singer}.} \bibinfo{year}{2011}\natexlab{}.
\newblock \showarticletitle{Adaptive subgradient methods for online learning
  and stochastic optimization}.
\newblock \bibinfo{journal}{\emph{Journal of machine learning research}}
  \bibinfo{volume}{12}, \bibinfo{number}{Jul} (\bibinfo{year}{2011}),
  \bibinfo{pages}{2121--2159}.
\newblock


\bibitem[\protect\citeauthoryear{Goldberg and Levy}{Goldberg and Levy}{2014}]%
        {goldberg2014word2vec}
\bibfield{author}{\bibinfo{person}{Yoav Goldberg} {and} \bibinfo{person}{Omer
  Levy}.} \bibinfo{year}{2014}\natexlab{}.
\newblock \showarticletitle{word2vec Explained: deriving Mikolov et al.'s
  negative-sampling word-embedding method}.
\newblock \bibinfo{journal}{\emph{arXiv preprint arXiv:1402.3722}}
  (\bibinfo{year}{2014}).
\newblock


\bibitem[\protect\citeauthoryear{Graves, Mohamed, and Hinton}{Graves
  et~al\mbox{.}}{2013}]%
        {graves2013speech}
\bibfield{author}{\bibinfo{person}{Alex Graves}, \bibinfo{person}{Abdel-rahman
  Mohamed}, {and} \bibinfo{person}{Geoffrey Hinton}.}
  \bibinfo{year}{2013}\natexlab{}.
\newblock \showarticletitle{Speech recognition with deep recurrent neural
  networks}. In \bibinfo{booktitle}{\emph{2013 IEEE international conference on
  acoustics, speech and signal processing}}. IEEE, \bibinfo{pages}{6645--6649}.
\newblock


\bibitem[\protect\citeauthoryear{He, Zhang, Ren, and Sun}{He
  et~al\mbox{.}}{2016}]%
        {he2016deep}
\bibfield{author}{\bibinfo{person}{Kaiming He}, \bibinfo{person}{Xiangyu
  Zhang}, \bibinfo{person}{Shaoqing Ren}, {and} \bibinfo{person}{Jian Sun}.}
  \bibinfo{year}{2016}\natexlab{}.
\newblock \showarticletitle{Deep residual learning for image recognition}. In
  \bibinfo{booktitle}{\emph{Proceedings of the IEEE conference on computer
  vision and pattern recognition}}. \bibinfo{pages}{770--778}.
\newblock


\bibitem[\protect\citeauthoryear{Hinton, Deng, Yu, Dahl, Mohamed, Jaitly,
  Senior, Vanhoucke, Nguyen, Sainath, et~al\mbox{.}}{Hinton
  et~al\mbox{.}}{2012}]%
        {hinton2012deep}
\bibfield{author}{\bibinfo{person}{Geoffrey Hinton}, \bibinfo{person}{Li Deng},
  \bibinfo{person}{Dong Yu}, \bibinfo{person}{George~E Dahl},
  \bibinfo{person}{Abdel-rahman Mohamed}, \bibinfo{person}{Navdeep Jaitly},
  \bibinfo{person}{Andrew Senior}, \bibinfo{person}{Vincent Vanhoucke},
  \bibinfo{person}{Patrick Nguyen}, \bibinfo{person}{Tara~N Sainath},
  {et~al\mbox{.}}} \bibinfo{year}{2012}\natexlab{}.
\newblock \showarticletitle{Deep neural networks for acoustic modeling in
  speech recognition: The shared views of four research groups}.
\newblock \bibinfo{journal}{\emph{IEEE Signal processing magazine}}
  \bibinfo{volume}{29}, \bibinfo{number}{6} (\bibinfo{year}{2012}),
  \bibinfo{pages}{82--97}.
\newblock


\bibitem[\protect\citeauthoryear{Hornik}{Hornik}{1991}]%
        {hornik1991approximation}
\bibfield{author}{\bibinfo{person}{Kurt Hornik}.}
  \bibinfo{year}{1991}\natexlab{}.
\newblock \showarticletitle{Approximation capabilities of multilayer
  feedforward networks}.
\newblock \bibinfo{journal}{\emph{Neural networks}} \bibinfo{volume}{4},
  \bibinfo{number}{2} (\bibinfo{year}{1991}), \bibinfo{pages}{251--257}.
\newblock


\bibitem[\protect\citeauthoryear{Huang, Liu, Van Der~Maaten, and
  Weinberger}{Huang et~al\mbox{.}}{2017}]%
        {huang2017densely}
\bibfield{author}{\bibinfo{person}{Gao Huang}, \bibinfo{person}{Zhuang Liu},
  \bibinfo{person}{Laurens Van Der~Maaten}, {and} \bibinfo{person}{Kilian~Q
  Weinberger}.} \bibinfo{year}{2017}\natexlab{}.
\newblock \showarticletitle{Densely connected convolutional networks}. In
  \bibinfo{booktitle}{\emph{Proceedings of the IEEE conference on computer
  vision and pattern recognition}}. \bibinfo{pages}{4700--4708}.
\newblock


\bibitem[\protect\citeauthoryear{Kingma and Ba}{Kingma and Ba}{2015}]%
        {kingma2014adam}
\bibfield{author}{\bibinfo{person}{Diederik~P Kingma} {and}
  \bibinfo{person}{Jimmy Ba}.} \bibinfo{year}{2015}\natexlab{}.
\newblock \showarticletitle{Adam: A method for stochastic optimization}. In
  \bibinfo{booktitle}{\emph{International Conference on Learning
  Representations 2015}}.
\newblock


\bibitem[\protect\citeauthoryear{Kipf and Welling}{Kipf and Welling}{2016}]%
        {kipf2016semi}
\bibfield{author}{\bibinfo{person}{Thomas~N Kipf} {and} \bibinfo{person}{Max
  Welling}.} \bibinfo{year}{2016}\natexlab{}.
\newblock \showarticletitle{Semi-supervised classification with graph
  convolutional networks}.
\newblock \bibinfo{journal}{\emph{arXiv preprint arXiv:1609.02907}}
  (\bibinfo{year}{2016}).
\newblock


\bibitem[\protect\citeauthoryear{Krizhevsky and Hinton}{Krizhevsky and
  Hinton}{2009}]%
        {krizhevsky2009learning}
\bibfield{author}{\bibinfo{person}{Alex Krizhevsky} {and}
  \bibinfo{person}{Geoffrey Hinton}.} \bibinfo{year}{2009}\natexlab{}.
\newblock \bibinfo{booktitle}{\emph{Learning multiple layers of features from
  tiny images}}.
\newblock \bibinfo{type}{{T}echnical {R}eport}.
\newblock


\bibitem[\protect\citeauthoryear{Krizhevsky, Sutskever, and Hinton}{Krizhevsky
  et~al\mbox{.}}{2012}]%
        {krizhevsky2012imagenet}
\bibfield{author}{\bibinfo{person}{Alex Krizhevsky}, \bibinfo{person}{Ilya
  Sutskever}, {and} \bibinfo{person}{Geoffrey~E Hinton}.}
  \bibinfo{year}{2012}\natexlab{}.
\newblock \showarticletitle{Imagenet classification with deep convolutional
  neural networks}. In \bibinfo{booktitle}{\emph{Advances in neural information
  processing systems}}. \bibinfo{pages}{1097--1105}.
\newblock


\bibitem[\protect\citeauthoryear{Lan}{Lan}{2012}]%
        {lan2012optimal}
\bibfield{author}{\bibinfo{person}{Guanghui Lan}.}
  \bibinfo{year}{2012}\natexlab{}.
\newblock \showarticletitle{An optimal method for stochastic composite
  optimization}.
\newblock \bibinfo{journal}{\emph{Mathematical Programming}}
  \bibinfo{volume}{133}, \bibinfo{number}{1-2} (\bibinfo{year}{2012}),
  \bibinfo{pages}{365--397}.
\newblock


\bibitem[\protect\citeauthoryear{LeCun, Boser, Denker, Henderson, Howard,
  Hubbard, and Jackel}{LeCun et~al\mbox{.}}{1989}]%
        {lecun1989backpropagation}
\bibfield{author}{\bibinfo{person}{Yann LeCun}, \bibinfo{person}{Bernhard
  Boser}, \bibinfo{person}{John~S Denker}, \bibinfo{person}{Donnie Henderson},
  \bibinfo{person}{Richard~E Howard}, \bibinfo{person}{Wayne Hubbard}, {and}
  \bibinfo{person}{Lawrence~D Jackel}.} \bibinfo{year}{1989}\natexlab{}.
\newblock \showarticletitle{Backpropagation applied to handwritten zip code
  recognition}.
\newblock \bibinfo{journal}{\emph{Neural computation}} \bibinfo{volume}{1},
  \bibinfo{number}{4} (\bibinfo{year}{1989}), \bibinfo{pages}{541--551}.
\newblock


\bibitem[\protect\citeauthoryear{LeCun, Bottou, Bengio, and Haffner}{LeCun
  et~al\mbox{.}}{1998}]%
        {lecun1998gradient}
\bibfield{author}{\bibinfo{person}{Yann LeCun}, \bibinfo{person}{L{\'e}on
  Bottou}, \bibinfo{person}{Yoshua Bengio}, {and} \bibinfo{person}{Patrick
  Haffner}.} \bibinfo{year}{1998}\natexlab{}.
\newblock \showarticletitle{Gradient-based learning applied to document
  recognition}.
\newblock \bibinfo{journal}{\emph{Proc. IEEE}} \bibinfo{volume}{86},
  \bibinfo{number}{11} (\bibinfo{year}{1998}), \bibinfo{pages}{2278--2324}.
\newblock


\bibitem[\protect\citeauthoryear{Li, Chen, and Shen}{Li et~al\mbox{.}}{2019}]%
        {li2019stereo}
\bibfield{author}{\bibinfo{person}{Peiliang Li}, \bibinfo{person}{Xiaozhi
  Chen}, {and} \bibinfo{person}{Shaojie Shen}.}
  \bibinfo{year}{2019}\natexlab{}.
\newblock \showarticletitle{Stereo r-cnn based 3d object detection for
  autonomous driving}. In \bibinfo{booktitle}{\emph{Proceedings of the IEEE
  Conference on Computer Vision and Pattern Recognition}}.
  \bibinfo{pages}{7644--7652}.
\newblock


\bibitem[\protect\citeauthoryear{Mnih, Kavukcuoglu, Silver, Graves, Antonoglou,
  Wierstra, and Riedmiller}{Mnih et~al\mbox{.}}{2013}]%
        {mnih2013playing}
\bibfield{author}{\bibinfo{person}{Volodymyr Mnih}, \bibinfo{person}{Koray
  Kavukcuoglu}, \bibinfo{person}{David Silver}, \bibinfo{person}{Alex Graves},
  \bibinfo{person}{Ioannis Antonoglou}, \bibinfo{person}{Daan Wierstra}, {and}
  \bibinfo{person}{Martin Riedmiller}.} \bibinfo{year}{2013}\natexlab{}.
\newblock \showarticletitle{Playing atari with deep reinforcement learning}.
\newblock \bibinfo{journal}{\emph{arXiv preprint arXiv:1312.5602}}
  (\bibinfo{year}{2013}).
\newblock


\bibitem[\protect\citeauthoryear{Mnih, Kavukcuoglu, Silver, Rusu, Veness,
  Bellemare, Graves, Riedmiller, Fidjeland, Ostrovski, et~al\mbox{.}}{Mnih
  et~al\mbox{.}}{2015}]%
        {mnih2015human}
\bibfield{author}{\bibinfo{person}{Volodymyr Mnih}, \bibinfo{person}{Koray
  Kavukcuoglu}, \bibinfo{person}{David Silver}, \bibinfo{person}{Andrei~A
  Rusu}, \bibinfo{person}{Joel Veness}, \bibinfo{person}{Marc~G Bellemare},
  \bibinfo{person}{Alex Graves}, \bibinfo{person}{Martin Riedmiller},
  \bibinfo{person}{Andreas~K Fidjeland}, \bibinfo{person}{Georg Ostrovski},
  {et~al\mbox{.}}} \bibinfo{year}{2015}\natexlab{}.
\newblock \showarticletitle{Human-level control through deep reinforcement
  learning}.
\newblock \bibinfo{journal}{\emph{Nature}} \bibinfo{volume}{518},
  \bibinfo{number}{7540} (\bibinfo{year}{2015}), \bibinfo{pages}{529--533}.
\newblock


\bibitem[\protect\citeauthoryear{Nesterov}{Nesterov}{1983}]%
        {nesterov27method}
\bibfield{author}{\bibinfo{person}{Yu Nesterov}.}
  \bibinfo{year}{1983}\natexlab{}.
\newblock \showarticletitle{A method of solving a convex programming problem
  with convergence rate $O (1/k^2)$}. In \bibinfo{booktitle}{\emph{Soviet
  Mathematics Doklady}}, Vol.~\bibinfo{volume}{27}.
  \bibinfo{pages}{372--–376}.
\newblock


\bibitem[\protect\citeauthoryear{Ramchoun, Idrissi, Ghanou, and
  Ettaouil}{Ramchoun et~al\mbox{.}}{2016}]%
        {ramchoun2016multilayer}
\bibfield{author}{\bibinfo{person}{Hassan Ramchoun}, \bibinfo{person}{Mohammed
  Amine~Janati Idrissi}, \bibinfo{person}{Youssef Ghanou}, {and}
  \bibinfo{person}{Mohamed Ettaouil}.} \bibinfo{year}{2016}\natexlab{}.
\newblock \showarticletitle{Multilayer Perceptron: Architecture Optimization
  and Training.}
\newblock \bibinfo{journal}{\emph{International Journal of Interactive
  Multimedia and Artificial Intelligence}} \bibinfo{volume}{4},
  \bibinfo{number}{1} (\bibinfo{year}{2016}), \bibinfo{pages}{26--30}.
\newblock


\bibitem[\protect\citeauthoryear{Robbins and Monro}{Robbins and Monro}{1951}]%
        {robbins1951stochastic}
\bibfield{author}{\bibinfo{person}{Herbert Robbins} {and}
  \bibinfo{person}{Sutton Monro}.} \bibinfo{year}{1951}\natexlab{}.
\newblock \showarticletitle{A stochastic approximation method}.
\newblock \bibinfo{journal}{\emph{The annals of mathematical statistics}}
  (\bibinfo{year}{1951}), \bibinfo{pages}{400--407}.
\newblock


\bibitem[\protect\citeauthoryear{Rumelhart, Durbin, Golden, and
  Chauvin}{Rumelhart et~al\mbox{.}}{1995}]%
        {rumelhart1995backpropagation}
\bibfield{author}{\bibinfo{person}{David~E Rumelhart}, \bibinfo{person}{Richard
  Durbin}, \bibinfo{person}{Richard Golden}, {and} \bibinfo{person}{Yves
  Chauvin}.} \bibinfo{year}{1995}\natexlab{}.
\newblock \showarticletitle{Backpropagation: The basic theory}.
\newblock \bibinfo{journal}{\emph{Backpropagation: Theory, architectures and
  applications}} (\bibinfo{year}{1995}), \bibinfo{pages}{1--34}.
\newblock


\bibitem[\protect\citeauthoryear{Rumelhart, Hinton, and Williams}{Rumelhart
  et~al\mbox{.}}{1986}]%
        {rumelhart1986learning}
\bibfield{author}{\bibinfo{person}{David~E Rumelhart},
  \bibinfo{person}{Geoffrey~E Hinton}, {and} \bibinfo{person}{Ronald~J
  Williams}.} \bibinfo{year}{1986}\natexlab{}.
\newblock \showarticletitle{Learning representations by back-propagating
  errors}.
\newblock \bibinfo{journal}{\emph{Nature}} \bibinfo{volume}{323},
  \bibinfo{number}{6088} (\bibinfo{year}{1986}), \bibinfo{pages}{533--536}.
\newblock


\bibitem[\protect\citeauthoryear{Tieleman and Hinton}{Tieleman and
  Hinton}{2012}]%
        {tieleman2012lecture}
\bibfield{author}{\bibinfo{person}{Tijmen Tieleman} {and}
  \bibinfo{person}{Geoffrey Hinton}.} \bibinfo{year}{2012}\natexlab{}.
\newblock \bibinfo{booktitle}{\emph{RMSProp, COURSERA: Neural networks for
  machine learning}}.
\newblock \bibinfo{type}{{T}echnical {R}eport}.
\newblock


\bibitem[\protect\citeauthoryear{Tseng}{Tseng}{1998}]%
        {tseng1998incremental}
\bibfield{author}{\bibinfo{person}{Paul Tseng}.}
  \bibinfo{year}{1998}\natexlab{}.
\newblock \showarticletitle{An incremental gradient (-projection) method with
  momentum term and adaptive stepsize rule}.
\newblock \bibinfo{journal}{\emph{SIAM Journal on Optimization}}
  \bibinfo{volume}{8}, \bibinfo{number}{2} (\bibinfo{year}{1998}),
  \bibinfo{pages}{506--531}.
\newblock


\bibitem[\protect\citeauthoryear{Wen, Zhang, Li, and Qiao}{Wen
  et~al\mbox{.}}{2016}]%
        {wen2016discriminative}
\bibfield{author}{\bibinfo{person}{Yandong Wen}, \bibinfo{person}{Kaipeng
  Zhang}, \bibinfo{person}{Zhifeng Li}, {and} \bibinfo{person}{Yu Qiao}.}
  \bibinfo{year}{2016}\natexlab{}.
\newblock \showarticletitle{A discriminative feature learning approach for deep
  face recognition}. In \bibinfo{booktitle}{\emph{European conference on
  computer vision}}. Springer, \bibinfo{pages}{499--515}.
\newblock


\bibitem[\protect\citeauthoryear{Xiong, Droppo, Huang, Seide, Seltzer, Stolcke,
  Yu, and Zweig}{Xiong et~al\mbox{.}}{2016}]%
        {xiong2016achieving}
\bibfield{author}{\bibinfo{person}{Wayne Xiong}, \bibinfo{person}{Jasha
  Droppo}, \bibinfo{person}{Xuedong Huang}, \bibinfo{person}{Frank Seide},
  \bibinfo{person}{Mike Seltzer}, \bibinfo{person}{Andreas Stolcke},
  \bibinfo{person}{Dong Yu}, {and} \bibinfo{person}{Geoffrey Zweig}.}
  \bibinfo{year}{2016}\natexlab{}.
\newblock \showarticletitle{Achieving human parity in conversational speech
  recognition}.
\newblock \bibinfo{journal}{\emph{arXiv preprint arXiv:1610.05256}}
  (\bibinfo{year}{2016}).
\newblock


\bibitem[\protect\citeauthoryear{Zeiler}{Zeiler}{2012}]%
        {zeiler2012adadelta}
\bibfield{author}{\bibinfo{person}{Matthew~D Zeiler}.}
  \bibinfo{year}{2012}\natexlab{}.
\newblock \showarticletitle{Adadelta: an adaptive learning rate method}.
\newblock \bibinfo{journal}{\emph{arXiv preprint arXiv:1212.5701}}
  (\bibinfo{year}{2012}).
\newblock


\end{thebibliography}


\end{document}